\documentclass[runningheads]{llncs}

\usepackage{booktabs} 
\usepackage{colortbl} 
\usepackage{graphicx} 
\usepackage[hidelinks]{hyperref}
\usepackage{amsmath}
\usepackage{amssymb}
\usepackage{array}
\usepackage[dvipsnames, svgnames, x11names]{xcolor}
\usepackage{geometry}
\begin{document}
%
\titlerunning{A high-level vision task-driven image fusion network via domain transformation}
\title{HSFusion: A high-level vision task-driven infrared and visible image fusion network via semantic and geometric domain transformation}

\author{Chengjie Jiang\inst{1} \and Xiaowen Liu\inst{2} \and Bowen Zheng\inst{3} \and Lu Bai\inst{4}\and Jing Li\inst{5}\thanks{Corresponding author: lijing2017@cufe.edu.cn}}
\authorrunning{F. Author et al.}
\institute{Central University of Finance and Economics \and People's Public Security University of China \and Beijing Normal University}

\maketitle              
\begin{abstract}
Infrared and visible image fusion has been developed from vision perception oriented fusion methods to strategies which both consider the vision perception and high-level vision task. However, the existing task-driven methods fail to address the domain gap between semantic and geometric representation. To overcome these issues, we propose a high-level vision task-driven infrared and visible image fusion network via semantic and geometric domain transformation, terms as HSFusion. Specifically, to minimize the gap between semantic and geometric representation, we design two separate domain transformation branches by CycleGAN framework,  and each includes two processes: the forward segmentation process and the reverse reconstruction process. CycleGAN is capable of learning domain transformation patterns, and the reconstruction process of CycleGAN is conducted under the constraint of these patterns. Thus, our method can  significantly facilitate the integration of semantic and geometric information and further reduces the domain gap. In fusion stage, we integrate the infrared and visible features that extracted from the reconstruction process of two seperate CycleGANs to obtain the fused result. These features, containing varying proportions of semantic and geometric information, can significantly enhance the high level vision tasks. Additionally, we generate masks based on segmentation results to guide the fusion task. These masks can provide semantic priors, and we design adaptive weights for two distinct areas in the masks to facilitate image fusion. Finally, we conducted comparative experiments between our method and eleven other state-of-the-art methods, demonstrating that our approach surpasses others in both visual appeal and semantic segmentation task.

\keywords{Infrared  \and Visible \and Image fusion \and Semantic segmentation.}
\end{abstract}

\begin{figure}
\includegraphics[width=\textwidth]{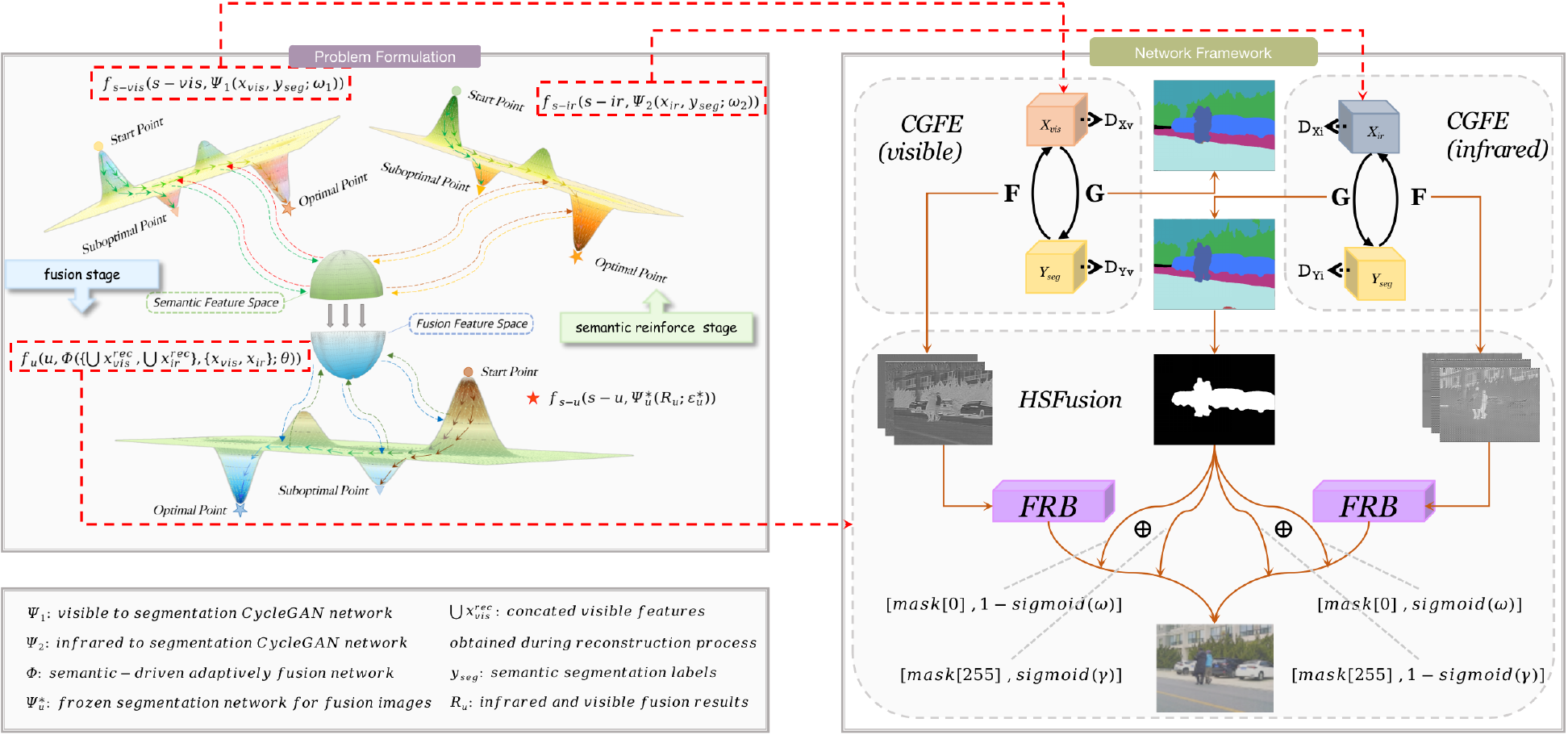}
\caption{The problem formulation and network. The left part illustrates the interaction between image fusion and segmentation tasks in the latent space. The right part displays the simplified network architecture of HSFusion.} \label{fig:fig1}
\end{figure}

\section{Introduction}
\label{sec:introduction}
The different sensors capture information from varying perspectives. For instance, visible sensors can clearly capture an object's texture details, however they are susceptible to extreme conditions such as darkness, strong light, or obstructions caused by rain and fog. In contrast, infrared sensor, which captures the information by thermal radiation, is adept at capturing the contours of objects and robust to environmental variations. Nevertheless, the lack of detailed information limits their ability in high-level vision tasks such as semantic segmentation. Image fusion aims to integrate the latent representations of source images from different modalities into a single fused image, which contains more comprehensive and complementary information to satisfy both visual perception and high-level vision tasks~\cite{ref_lncs1}.

From a methodological perspective, infrared and visible fusion algorithms can be categorized into traditional algorithms and deep learning-based methods. Traditional methods often require manually formulated fusion strategies, which lead to complexity, low generalization, and a lack of end-to-end learning capabilities. Therefore, deep learning-based methods have been applied into this task due to their excellent feature extraction abilities and capacity of end-to-end training~\cite{ref_lncs2}. The deep learning-based methods include Autoencoders (AE)~\cite{ref_lncs3}, Convolutional Neural Networks (CNN)~\cite{ref_lncs4}, or Generative Adversarial Networks (GAN)~\cite{ref_lncs5}. However, most of early methods pay more attention to local features by CNN, while have difficulty in representing long-range features. Thus, the transformer-based methods are introduced to address this problem~\cite{ref_lncs6}~\cite{ref_lncs7}. Although the aforementioned methods have achieved commendable visual perception in fused images, they ignore the relationship between image fusion and downstream high-level vision tasks, which limits the application of image fusion.

Therefore, to integrate image fusion with high-level vision tasks such as semantic segmentation, several methods have made attempts from three dimensions. Tang et al. designs a segmentation-fusion cascaded network with semantic segmentation-related loss functions~\cite{ref_lncs8}. Wu et al. takes a pre-trained segmentation network to obtain segmentation results, which then guide the fusion task~\cite{ref_lncs9}. Additionally, Tang et al. incorporates a semantic information injection module during the fusion process, achieving the infusion of semantics at the feature level~\cite{ref_lncs10}. However, due to the divergent optimization directions of segmentation and fusion tasks, the simple cascaded structure is insufficient to constrain the optimization trends of these distinct tasks. Furthermore, due to the heterogeneity between semantic and geometric features, injecting semantic information without proper constraints impedes their seamless integration.

To address the above issues, we propose a high-level vision task-driven infrared and visible image fusion network via semantic and geometric domain transformation, which jointly optimizes fusion and segmentation tasks, terms as HSFusion. Our method comprises two separate pretrained feature extractors, along with an adaptive feature fusion network guided by the semantic segmentation result, which is presented in Fig.~\ref{fig:fig1}. Specifically, considering the inherent properties of infrared and visible images, we utilize two separated CycleGANs~\cite{ref_proc1} with non-shared weights as feature extractors(CGFE). Each CycleGAN framework encompasses two processes: the forward segmentation process, which transforms source images into semantic segmentation results, and the reverse reconstruction process, which aims to reconstruct source images from semantic segmentation results. CGFE is adept at learning stable transformation patterns between the infrared/visible and semantic domains. Under these domain transformation pattern constraints, the methodology ensures that the reconstructed infrared/visible features can minimize the gap between semantic and geometric information. In the fusion stage, we meticulously select thermal objects(such as traffic lights, person, car, etc.) based on the previous obtained segmentation results to create masks with fine granularity. Simultaneously, we use the features obtained during the reconstruction process as inputs. By designing adaptive weights, this approach assigns greater emphasis to infrared features in thermal source regions, while visible features in non-thermal regions. Thus, the adaptive feature weighting strategy takes the strengths of each modality's features into consideration, and enhance the interpretability of the fusion process.

The main contributions of our research can be summarized as follows:

\begin{enumerate}
    \item 
    To better serve high-level vision tasks, we establish two independent pretrained feature extractions to fully extract both semantic and geometric information of infrared and visible images, which aims to enhance not only the visual perception but also the semantic representation of the fusion results.
    \item 
    To minimize the gap between semantic and geometric domains, we employ CycleGAN structure to learn the latent transformation patterns of different domains. Then, we can achieve the integration of semantic and geometric information under the constraint of domain transformation patterns.
    \item 
    We utilize semantic masks that generated by two pretrained schemes to guide the image fusion process, which can enhance the complementary semantic prior of different source images and further improve the performance of fusion and high-level vision tasks, simultaneously.
    \item 
    Experiments demonstrate qualitatively and quantitatively that our HSFusion achieves state-of-the-art results both in visual perception and high-level semantic segmentation task.
\end{enumerate}

The remainder of this paper is organized as follows. Section~\ref{sec:related-works}  introduces the related works about visual perception-driven fusion methods and high-level vision task-driven fusion methods. Section~\ref{sec:methodology} shows the details of our methodology. In Section~\ref{sec:experimental validation}, we present the fused results of our method and the compared methods on the public dataset, followed by some concluding remarks in Section~\ref{sec:conclusion}.

\section{Related Works}
\label{sec:related-works}
Although previous deep learning-based methods have achieved impressive results in visual perception, they seldom consider the applications in high-level vision tasks. Therefore, high-level vision task-driven fusion methods have been proposed. In this section, we will present the related works about visual perception-driven methods and the high-level vision task-driven methods.

\subsection{visual perception-driven methods}
\label{subsec:visual}
Deep learning-based image fusion methods have achieved remarkable performance in enhancing visual perception,  which include AE-, CNN-, GAN-, Transformer-based methods and other methods. AE-based methods typically involve a pretrained autoencoder for feature extraction and image reconstruction. However, this architecture is quite sensitive to the quality and diversity of the training data, which leads to them struggle to extract complex features. Therefore, Li et al. proposed the DenseBlocks and dense connections to improve the robustness of the feature extraction ~\cite{ref_lncs3}. Besides, the CNN-based methods were proposed by utilizing neural networks to construct weight maps or perform feature extraction~\cite{ref_lncs4}. Li et al. introduced the RFN-Nest architecture, which employs a combination of convolutional layers and max-pooling layers for effective down-sampling, facilitating the extraction of multi-scale features~\cite{ref_lncs11}.  The inherent unsupervised ability of GAN renders it particularly apt for image fusion task which is lack of ground truth. Ma et al. first introduced GAN model into image fusion, employing a single discriminator to force the fused images keep rich texture information~\cite{ref_lncs5}, and then, Ma et al. proposed a dual-discriminator conditional GAN to address the balance issues in infrared and visible image fusion task~\cite{ref_lncs13}. Meanwhile, Li et al. incorporated attention mechanisms into GAN-based image fusion methods~\cite{ref_lncs14}~\cite{ref_lncs15}.  In recent years, transformer  has also been introduced to image fusion task. Ma et al. designed domain-specific fusion units and domain-crossing fusion units, achieving comprehensive integration and global interaction of complementary information~\cite{ref_lncs6}. Chang et al. implemented image fusion via transformer architecture to extract global features ~\cite{ref_lncs16}. Considering the Transformer's performance in global modeling and the CNN's superiority in local feature extraction, Zhao et al. proposed a Transformer-CNN feature extractor, which effectively captures both low-frequency global features and high-frequency local information~\cite{ref_proc2}.

\subsection{high-level vision task-driven methods}
\label{subsec:high-level}
To enhance the effectiveness of fusion results in high-level vision tasks, Tang et al. proposed a semantic-driven image fusion algorithm~\cite{ref_lncs17}, which involves designing semantic-related loss functions and employing backpropagation to achieve a joint optimization. Additionally, Liu et al. designed a cascaded network that utilizes the results of image fusion to guide the salient object detection task~\cite{ref_proc3}. However, the simple joint training ignores the feature gap between two different-level tasks. To address this, Zhao et al. utilized meta-learning (MetaFusion) to reduce the feature gap to fuse infrared and visible images \cite{ref_proc4}. Furthermore, Ma et al. proposed progressive semantic injection and scene fidelity constraints to  enhance the performance of segmentation and fusion tasks ~\cite{ref_lncs10}. In addition, to address the issue of insufficient preservation and representation of semantic information in existing methods, Liu et al. developed a coupled network enhancing both pixel-level and semantic-level information~\cite{ref_proc5}. However, the current methods of semantic injection do not fundamentally address the gap between semantic and geometric spaces. Even with the use of scene fidelity constraints, these approaches still fail to resolve the issue that introducing information from different domains will disrupt the distribution of original information, thereby producing additional noise. 

To this end, we propose a high-level vision task-driven image fusion network via semantic and geometric domain transformation. More specifically, we employ CycleGAN as feature extractor and design adaptive fusion strategy based on the semantic mask. This approach not only bridges the domain gap through the integration of semantic and geometric information but also leverages semantic segmentation outcomes to enhance the fusion process with semantic priors.

\begin{figure}[h]
\includegraphics[width=\textwidth]{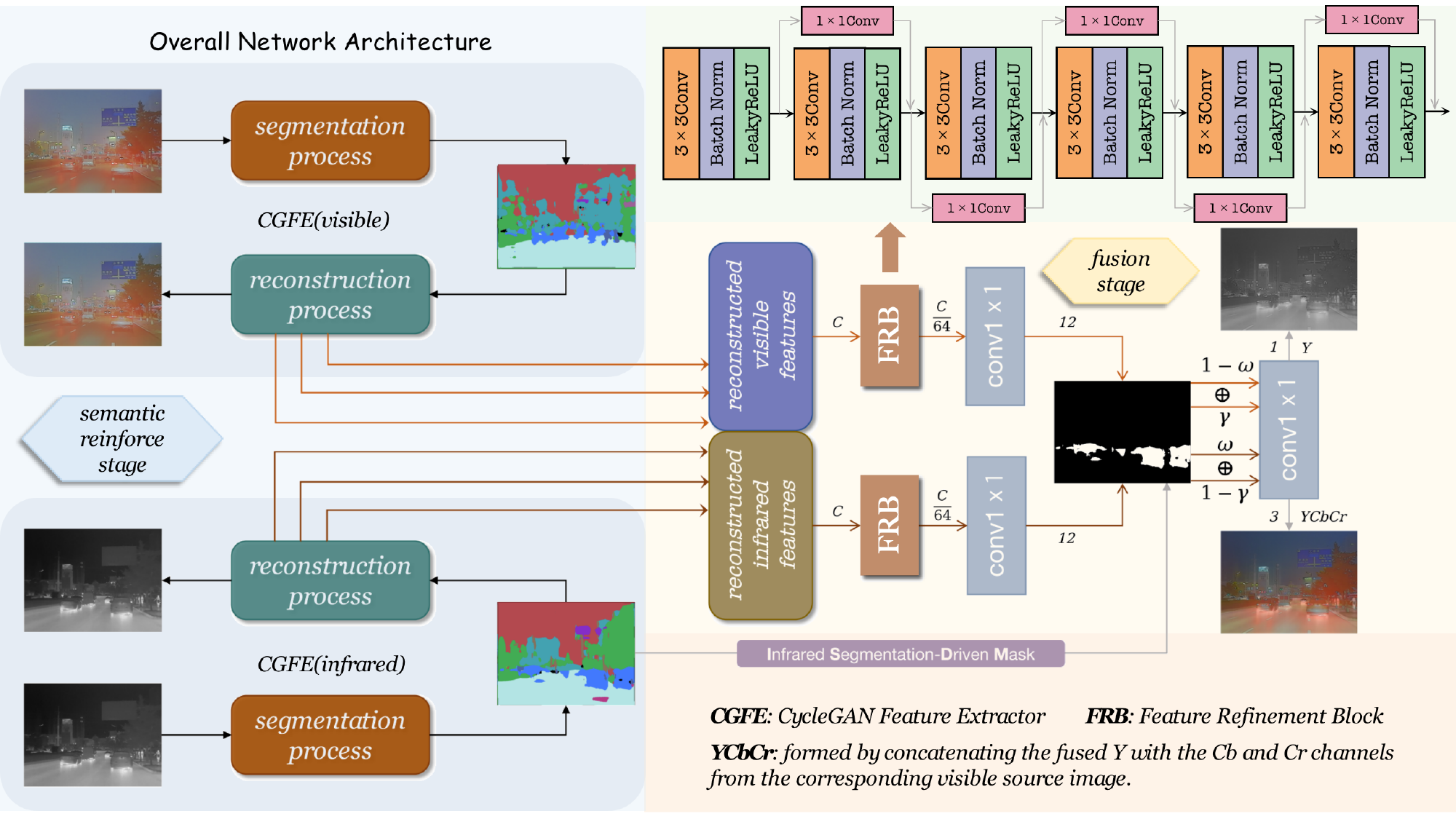}
\caption{The overall architecture of the proposed HSFusion. It comprises two main stages: the semantic reinforce stage and the fusion stage. The semantic reinforce stage encompasses two key processes: the segmentation process and the reconstruction process.} \label{fig:fig2}
\end{figure}

\section{Methodology}
\label{sec:methodology}
In this section, we discuss the problem formulation of jointly optimization between image fusion and semantic segmentation tasks, followed by the details of  architecture of our model and the loss functions.  

\subsection{Problem Formulation}
\label{subsec:problem formulation}
For most deep learning based methods, they tend to design a neural network that continuously optimizes the model by minimizing the loss relative to the ground truth. The optimization process can be formulated as:
\begin{equation}
\min_{\omega_k} f(N, \mathcal{N}_k(x, y; \omega_k)),
\end{equation}
where $N$ denotes the name of the task-related network $\mathcal{N}_k$ with learnable parameters $\omega_k$. $x$ and $y$ denotes the input of the network and the ground truth respectively. In semantic segmentation tasks, the ground truth $y$ denotes semantic segmentation labels $y_{seg}$.  In contrast, in fusion task, $y$ denotes the original infrared images $x_{ir}$ and visible images $x_{vis}$, which are used to calculate loss due to the lack of ground truth.

As shown in Fig.~\ref{fig:fig1}, the previous methods often treated fusion and segmentation as separate tasks. These approaches typically resulted in suboptimal performance in each individual task. However, we adeptly link the fusion and segmentation tasks within the latent space, thereby achieving optimal performance in both tasks. In our method, the entire architecture is divided into two stages: the semantic reinforce stage and the fusion stage. The optimization objective of the semantic reinforce stage can be formulated as:
\begin{equation}
\min_{\omega} f_{s}(s, \Psi(x, y_{\text{seg}}; \omega)),
\end{equation}
where $f_{s}(\cdot)$ denotes the segmentation network $\Psi$ using visible/infrared images $x$ as input. In the fusion stage, the optimization objective can be formulated as:
\begin{equation}
\min_{\theta} f_u \left( u, \Phi \left( \{ \bigcup x^{\text{rec}}_{\text{vis}}, \bigcup x^{\text{rec}}_{\text{ir}} \}, \{ x_{\text{vis}}, x_{\text{ir}} \}; \theta \right) \right),
\end{equation}
where $u$ is the name of the semantic-driven adaptive fusion network $\Phi$ with learnable parameters $\theta$. $\bigcup$ represents concatenation along the channel dimension, so that $\{ \bigcup x^{\text{rec}}_{\text{vis}}, \bigcup x^{\text{rec}}_{\text{ir}} \}$ denote the concatenated infrared and visible features, which obtained during the reconstruction process(we will discuss it detailly in Section~\ref{subsubsec:semantic reshaping phase}).

\begin{figure}[h]
\includegraphics[width=\textwidth]{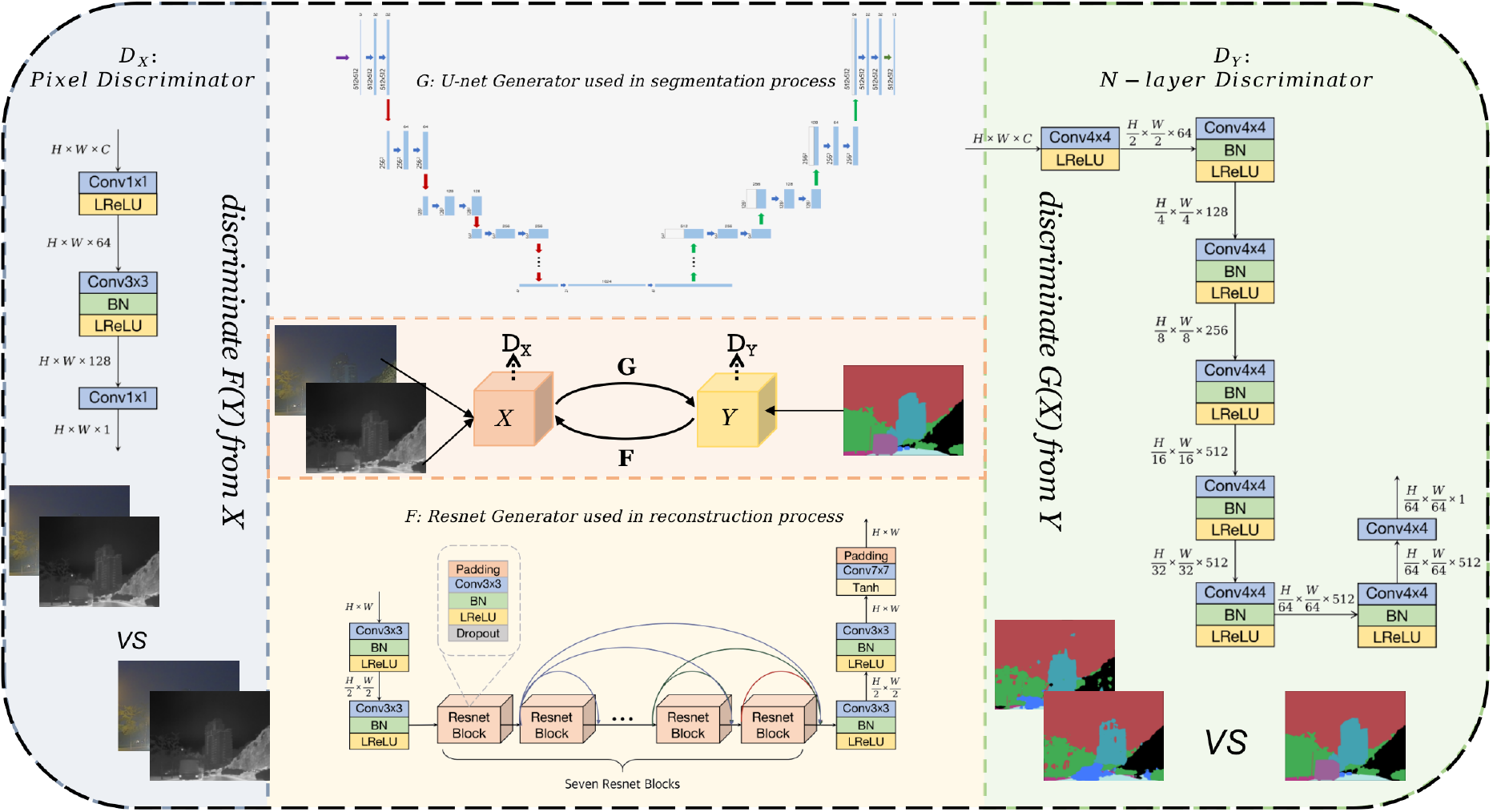}
\caption{The detailed architecture of the CGFE module. The central red section represents the overall architecture of the CGFE. The upper gray section is the details of generator $G$, which uses a U-net architecture. The lower yellow section outlines the architecture of generator $F$, and the left blue and right green sections respectively depict the detailed architectures of discriminators $D_x$ and $D_y$.} \label{fig:fig3}
\end{figure}

\subsection{Network Architecture}
\label{subsec:network architecture}
To enhance fusion and segmentation tasks simultaneously, we propose a high-level vision task-driven image fusion network via semantic and geometric domain transformation, which consists of two stages: the semantic reinforce stage and the fusion stage (as shown in Fig.~\ref{fig:fig2}). More specifically, the semantic reinforce stage encompasses two key processes: the segmentation process and reconstruction process, which executed by two separate CGFE. In addition, the fusion stage incorporates an ISDM (Infrared Segmentation-Driven Mask) module that leverages segmentation results to guide the fusion process, and to refine the features for the final reconstructed features, an FRB (Feature Refinement Block) module is designed in  fusion stage.

\subsubsection{Semantic Reinforce Stage.}
\label{subsubsec:semantic reshaping phase}
Given a pair of infrared image $I_{ir} \in \mathbb{R}^{H \times W \times 1}$ and visible image $I_{vis} \in \mathbb{R}^{H \times W \times 3}$, the segmentation process aims to obtain the semantic segmentation results $\mathcal{R}_{seg}^{vi/ir} \in \mathbb{R}^{H \times W \times n}$ for each source image, where $n$ denotes the number of segmentation categories. The $\mathcal{R}_{seg}^{vi/ir}$ need to approximate the real semantic segmentation labels $\mathcal{R}_{label} \in \mathbb{R}^{H \times W \times 1}$. Meanwhile, the reconstruction stage aims to get the reconstructed infrared image $\hat{I}_{ir} \in \mathbb{R}^{H \times W \times 1}$ or visible image $\hat{I}_{vis} \in \mathbb{R}^{H \times W \times 3}$. Considering the inherent properties of $I_{ir}$ and $I_{vis}$, the whole segmentation stage is based on two separate CGFE modules with non-shared parameters. The details of the CGFE model are presented in Fig.~\ref{fig:fig3}, which employs the CycleGAN structure, comprises a U-net Generator $G(\cdot)$~\cite{ref_lncs18} for the segmentation process, a Resnet Generator $F(\cdot)$ for the reconstruction process, along with a Pixel Discriminator $D_x(\cdot)$ and an $N$-layer Discriminator $D_y(\cdot)$.

For example, in the visible image branch, the U-net Generator $G(\cdot)$ is designed to learn a mapping $G: I_{\text{vis}} \rightarrow \mathcal{R}_{\text{label}}$, so that the distribution of $\mathcal{R}_{seg}^{vi} = G(I_{\text{vis}})$ can be indistinguishable from the distribution $\mathcal{R}_{\text{label}}$ using an adversarial loss. More specifically, $G(\cdot)$ employs the classic U-net architecture, including seven down-sampling steps followed by seven up-sampling steps with skip connections. This multi-scale transformation not only expands the receptive field, but also enables the extraction of both shallow and deep information. The skip connections reduce the loss of shallow geometric information while transitioning to the semantic domain.

Meanwhile, to construct the comprehensive domain transformation pattern, the Resnet Generator $F(\cdot)$ is introduced to learn an inverse mapping $F: \mathcal{R}_{\text{label}} \rightarrow I_{\text{vis}}$. Then, the reconstructed result $\hat{I}_{\text{vis}} = F(\mathcal{R}_{seg}^{vi})$ will be introduced in the cycle consistency loss to enforce $\hat{I}_{\text{vis}} \approx I_{\text{vis}}$. The details of $F(\cdot)$ are illustrated in the yellow region of Fig.~\ref{fig:fig3}: the input is conducted by two down-sampling operations, seven resnet blocks, two up-sampling operations, and then processed by an activation layer to complete the domain transformation, which can be formulated as:
\begin{equation}
\hat{I}_{\text{vis}} = \tanh(\uparrow^2(\text{Res}B^7(\downarrow^2(\mathcal{R}_{seg}^{vi})))),
\end{equation}
where \( \downarrow^n(\cdot) \) denotes down-sampling \( n \) times and \( \uparrow^n(\cdot) \) denotes up-sampling \( n \) times. \( \text{Res}B^n(\cdot) \) indicates \( n \) residual blocks, which is designed to integrate deep-level semantic information with shallow-level texture and structural information.

To ensure \( \mathcal{R}_{seg}^{vi} \) closely approximates \( \mathcal{R}_{\text{label}} \), the N-layer Discriminator \( D_y(\cdot) \), which utilizes the PatchGAN architecture~\cite{ref_proc6}, has been proposed. Considering semantic information often exhibits regional consistency and resides in deeper layers, \( D_y(\cdot) \) employs larger convolution kernels (4x4) for multiple down-sampling steps. The details of  \( D_y(\cdot) \) are shown in the right side of Fig.~\ref{fig:fig3}. Correspondingly, to ensure \( \hat{I}_{\text{vis}} \) closely resembles \( I_{\text{vis}} \), the Pixel Discriminator \( D_x(\cdot) \) has been introduced. It utilizes smaller convolution kernels and does not perform down-sampling, thereby achieving fine-grained discrimination at the pixel scale.

Therefore,  through the above conversion mechanism among the infrared/visible images and semantic representation, the CGFE module can learn the transformation patterns between the semantic and geometric domains. Constrained by these patterns, the array of features obtained during the reconstruction process, which contains varying proportions of semantic and geometric information, will consistently maintain a uniform distribution. So that the process of semantic injection will not introduce additional noise.

\subsubsection{Fusion Stage.}
\label{subsubsec:fusion phase}
In the fusion stage,  we take the  pretrained models ($\text{CGFE}_{\text{vis}}$ and $\text{CGFE}_{\text{ir}}$)  of  semantic reinforce stage as the backbone to extract the features of original images ( $I_{\text{vis}}$ and $I_{\text{ir}}$ ), which can be formulated as follow:
\begin{equation}
\left\{ \mathcal{R}^{\text{vi/ir}}_{\text{seg}}, \{F^i_{\text{vi/ir}} \, \text{where} \, i \in \{1,2,3,4,5\}\} \right\} = \text{CGFE}^*_{\text{vi/ir}}(I_{\text{vi/ir}}),
\end{equation}
where $F_{\text{vi}}^i$, $F_{\text{ir}}^i$ are the $i$-th elements within the set of features that generated in the reconstruction process. 

The details of the fusion stage are displayed in Fig.~\ref{fig:fig2}. Initially, the five features that extracted by the pretrained models ($\text{CGFE}_{\text{vis}}$ and $\text{CGFE}_{\text{ir}}$)  are up-sampled to the same spatial resolution, which can be formulated as follow:
\begin{equation}
F_{\text{vi/ir}}^{\text{rec}} = \bigcup_{i=0}^{4} \uparrow (F_{\text{vi/ir}}^i).
\end{equation}

Then, we take FRB modules to further refine these features. The specific details of the FRB are presented in Fig.~\ref{fig:fig2}. It reduces the number of channels to $1/64$ of the input by down-sampling operation to decrease the computational load for subsequent operations. Moreover, the FRB utilizes residual connections to prevent gradient vanishing. The refined infrared and visible features are calculated as:
\begin{equation}
F^{\text{ref}}_{\text{vi/ir}} = conv(FRB(F^{\text{rec}}_{\text{vi/ir}})).
\end{equation}

In addition, considering that infrared images contain sharp contour information, we employ a semantic-driven fusion approach and utilize an ISDM module to combine the advantages of different modalities.  More specially, the ISDM module aims to utilize thermal labels of the segmentation results $\mathcal{R}_{\text{seg}}^{\text{ir}}$ (such as Traffic Light, Person, Car, Truck, Bus, and Motorcycle) as the foreground, which are more prominent in infrared image, and the other objects are seen as background. The mask $M$ can be formulated as:
\begin{equation}
M(i,j) =
\begin{cases} 
0, & \text{if } R^{ir}_{seg}(i,j) \in \{4,8,9,10,11,12\} \\
255, & \text{otherwise}
\end{cases},
\end{equation}
where $\{4,8,9,10,11,12\}$ represents the segmentation labels of  traffic light, person, car, truck, bus, motorcycle, respectively, and the $(i,j)$ denotes the position of pixel. Then, as shown in Fig.~\ref{fig:fig2}, we weight these two regions independently by employing two adaptive weights. Thus, the fusion results can be calculated as:
\begin{align}
I_f &= \sigma(\omega) \times F_{\text{ir}}^{\text{ref}}(i, j| M(i, j) = 0) + (1-\sigma(\omega)) \times F_{\text{vi}}^{\text{ref}}(i, j| M(i, j) = 0) \nonumber\\
&\quad + (1-\sigma(\gamma)) \times F_{\text{ir}}^{\text{ref}}(k, l| M(k, l) = 255) + \sigma(\gamma) \times F_{\text{vi}}^{\text{ref}}(k, l| M(k, l) = 255),
\end{align}
where \( \omega \) and \( \gamma \) are the trainable parameters and $\sigma(\cdot)$ represents the $sigmoid(\cdot)$ function.

\subsection{Loss Function}
\label{loss function}
To improve the performance of fusion and segmentation tasks, we design distinct loss functions to constrain the semantic reinforce stage and the fusion stage. In the semantic reinforce stage, our loss functions include CycleGAN loss, semantic representation loss, and structural loss.  The total loss of the semantic reinforce stage $\mathcal{L}_{\text{sr}}$ can be formulated as follow: 
\begin{equation}
\begin{aligned}
\mathcal{L}_{\text{sr}}= \mathcal{L}_{\text{cg}}+ \lambda \mathcal{L}_{\text{sere}}+ \mathcal{L}_{\text{str}}
\end{aligned},
\end{equation}
where $\mathcal{L}_{\text{cg}}$ denotes CycleGAN loss, which is utilized to modify the whole domain transformation patterns. Moreover, we employ the Online hard example mining Cross-Entropy Loss (OhemCELoss)~\cite{OhemCELoss} to calculate the semantic representation loss $\mathcal{L}_{\text{sere}}$, which ensures the extraction of enough semantics during the segmentation process, and the $\lambda$ is the hyper-parameter.  $\mathcal{L}_{\text{str}}$ represents the structural loss calculated by the sum of SSIM loss~\cite{ref_lncs10} and Sobel loss~\cite{ref_proc5}, which can ensure that the reconstructed features contain rich texture and structural information.

In the fusion stage, we solely utilize geometric loss to calculate the fused image. The total loss $\mathcal{L}_{\text{geo}}$ in this stage consists of SSIM loss and MSE loss\cite{ref_lncs10}, which can be calculated as follow:

\begin{equation}
\mathcal{L}_{\text{geo}}=\mu \times (1 - \text{SSIM}(I_f, I_{\text{vis}})) + \rho \times \text{MSE}(I_f, I_{\text{vis}}) + \ \eta \times \text{MSE}(I_f, I_{\text{ir}}),
\end{equation}
where $\mu$, $\rho$, and $\eta$ are all the fixed hyper-parameters. 

\section{Experimental validation}
\label{sec:experimental validation}
In this section, we first present the details of experimental configurations and datasets. Then, we qualitatively and quantitatively compare our method with eleven state-of-the-art methods both on fusion and semantic segmentation tasks , and our method has the better performance in both two tasks.
\subsection{Experimental Details}
\subsubsection{Experimental Configurations.}
In our method, we train our fusion model by a two-stage training process. We first train the CGFE for 400 epochs to obtain multimodal features enriched with semantic information, then we train the fusion network with 400 epochs. In training process of our method, the fixed hyper-parameters are defined as follow:  $\lambda_{\text{sere}}=80$, $\mu=100$, $\rho=50$, $\eta=40$, meanwhile, the trainable parameters are initialed as: $\omega=0.5$, $\gamma=0.5$. In addition, we employ Adam to optimize our model, whose momentum parameter is set to 0.5 and the learning rate is initialized as $2 \times 10^{-4}$. Moreover, we train our model on the NVIDIA GeForce RTX 4090 with 24GB memory and AMD Ryzen 7950X with 4.5GHz.
\subsubsection{Experimental datasets.}
In our work, we train and test our method on FMB dataset, which contains image pairs of infrared image, visible image and semantic segmentation label~\cite{ref_url1}. In FMB, the training set includes 1220 image pairs and the test set contains 280 image pairs. For the qualitative analysis, we evaluate the fusion result by human perception. In addition, we use structural similarity index measure $(SSIM)$, correlation coefficient $(CC)$, peak signal-to-noise ratio $(PSNR)$, and modified fusion artifacts measure $(N^{AB/F})$ to compare the fusion methods in quantitative way.  Besides, to compare the performance of different methods on high-level vision task, we utilize the ViT-Adapter semantic segmentation model to test all the fused results~\cite{ref_proc9}. We calculate the mean Intersection over Union $(mIoU)$ to quantitatively compare the effectiveness of each fusion method in segmentation task.
\subsection{Fusion comparison and analysis}
\subsubsection{Qualitative comparison and analysis.}
\begin{figure}[t]
\includegraphics[width=\textwidth]{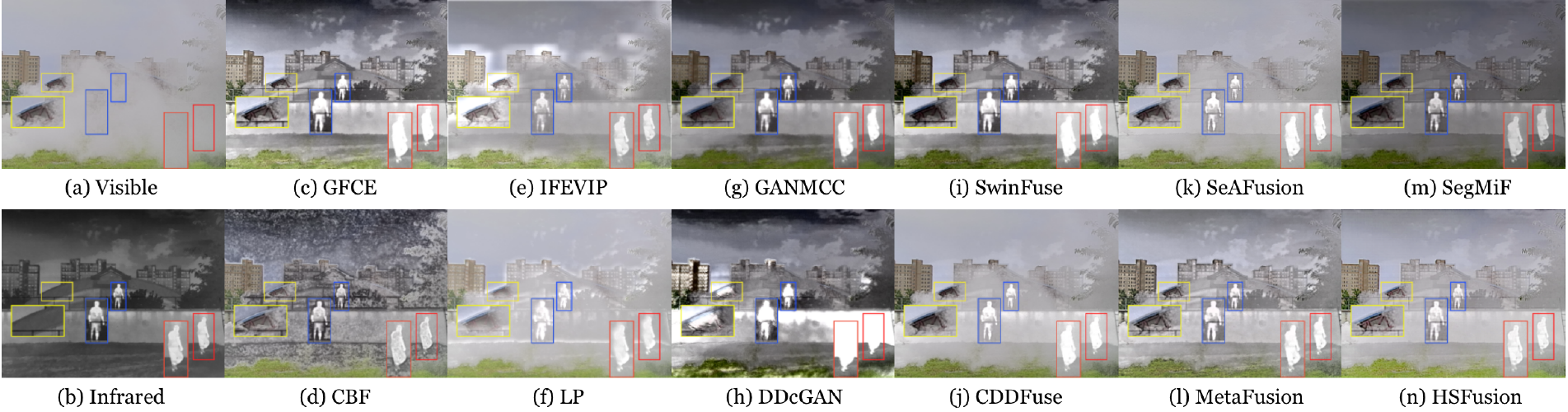}
\caption{Qualitative comparison of HSFusion with 11 state-of-the-art methods on the 00089 scene from the FMB dataset.} \label{fig:fig4}
\end{figure}

\begin{figure}[h]
\includegraphics[width=\textwidth]{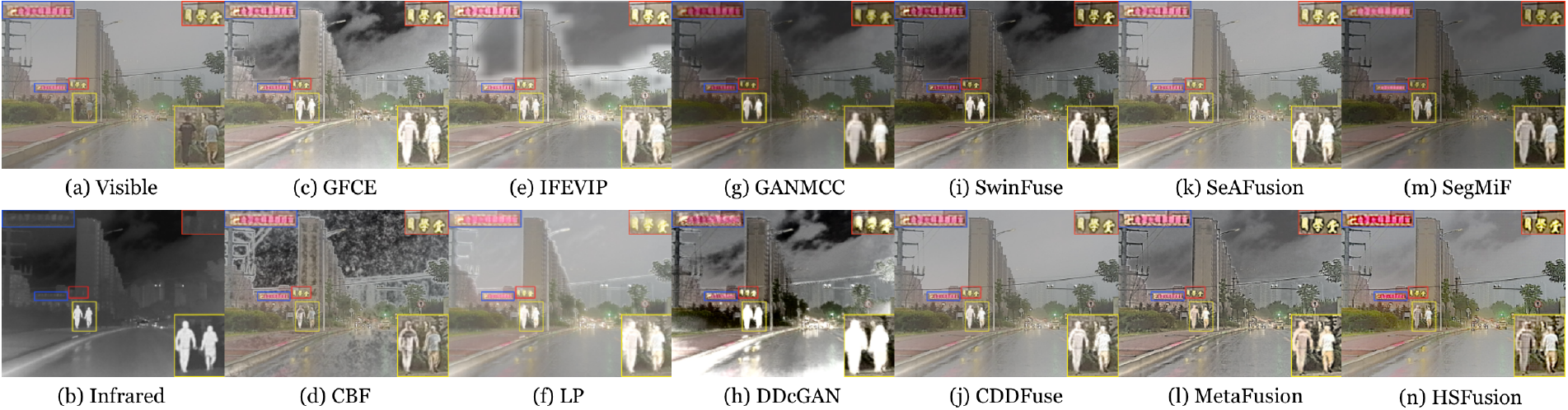}
\caption{Qualitative comparison of HSFusion with 11 state-of-the-art methods on the 00123 scene from the FMB dataset.} \label{fig:fig5}
\end{figure}

In comparsion experiments, we qualitatively and quantitatively compare our method with four traditional methods: GFCE~\cite{ref_lncs21}, CBF~\cite{ref_lncs22}, LP~\cite{ref_lncs23}, IFEVIP~\cite{ref_lncs24}; two GAN-based fusion methods: GANMCC~\cite{ref_lncs25}, DCcGAN~\cite{ref_lncs13}; two transformer-based methods: SwinFuse~\cite{ref_lncs26}, CDDFuse~\cite{ref_proc2}; and three high level vision task-driven methods: SeAFusion~\cite{ref_lncs8}, MetaFusion~\cite{ref_proc4}, SegMiF~\cite{ref_proc8}. The fused results are shown in Fig.~\ref{fig:fig4} and Fig.~\ref{fig:fig5}.

Fig.~\ref{fig:fig4} shows that our method preserves the information of people in the fog more clearly than two GAN-based methods. Moreover, our method also contains more background information than the other methods, where the fusion results from GFCE, CBF, IFEVIP, GANMCC, DCcGAN, SwinFuse, and SegMiF exhibit distortions in the background. Lastly, compared to all other methods, it can be observed that our method demonstrates the best clarity and color fidelity of the house structure in the yellow box.
\begin{table}
\centering
\caption{Quantitative comparison of eleven methods on the four metrics, which is performed on 280 pairs of images from the FMB dataset, where \textcolor{red}{red} values denote optimal, \textcolor{green!70!black}{green} values denote sub-optimal. Specifically, the last four methods are high-level vision task-driven methods.}
\label{fusion_result}
\resizebox{\textwidth}{!}{%
\begin{tabular}{lcccccccccccccc}
\toprule
& CBF & LP & IFEVIP & GFCE & DDcGAN & GANMcC & SwinFuse & CDDFuse & SeAFusion & SegMiF & MetaFusion & \textbf{HSFusion} \\
\midrule
Parameter(M) & - & - & - & - & 1.10 & 1.86 & 23.07 & 2.27 & 12.47 & 43.53 & 2.15 & 4.82\\
\midrule
$SSIMX$ $\uparrow$ & 1.0584 & 1.4587 & 1.4524 & 1.3479 & 1.1967 & \textcolor{green!70!black}{1.4731} & 1.3477 & 1.4468 & 1.4405 & 1.4596 & 1.3021 & \textcolor{red}{1.4886} \\
$CC$ $\uparrow$ & 0.4974 & 0.5765 & 0.5945 & 0.5542 & 0.5399 & \textcolor{green!70!black}{0.6467} & 0.6398 & 0.6303 & 0.6218 & 0.6394 & 0.6274 & \textcolor{red}{0.6569} \\
$PSNR$ $\uparrow$ & 15.8678 & 15.7677 & 15.1946 & 14.2826 & 12.6139 & 15.9299 & 14.9227 & \textcolor{green!70!black}{16.6050} & 16.4415 & 16.0861 & 15.5639 & \textcolor{red}{16.6501} \\
$N^{AB/F}$ $\downarrow$ & 0.0190 & \textcolor{green!70!black}{0.0055} & 0.0254 & 0.0644 & 0.0870 & 0.0058 & 0.0371 & 0.0344 & 0.0173 & 0.0158 & 0.1970 & \textcolor{red}{0.0046} \\
\bottomrule
\end{tabular}%
}
\end{table}
In addition, Fig.~\ref{fig:fig5} shows that our fusion results also has the superiority  than other methods in terms of fidelity background information, restoration of salient details, and clarity of texture. These advantages can be attributed to two main factors. Firstly, we adopted a semantic-driven adaptive weighting fusion strategy, which allows the fused images to capture enough salient information from infrared images while restore background information from visible image. Secondly, we employed various vision-related loss functions to further ensure the overall clarity and fidelity of the images.

\subsubsection{Quantitative comparison and analysis.}
The quantitative comparison results are illustrated in Table~\ref{fusion_result}, and our method has the best performance on $SSIM$, $CC$, $PSNR$ and $N^{AB/F}$ . The quantitative comparisons demonstrate that our method effectively preserves the structural and texture information of the source images, meanwhile, introduces minimal artifacts and noise during the fusion process. Moreover, compared with the high-level task-driven methods, our method also has fewer parameters than SeAFusion and SegMiF. As for MetaFusion, it has an even lower parameter count due to the meta-learning techniques and pretrained generators.
\begin{figure}
\includegraphics[width=\textwidth]{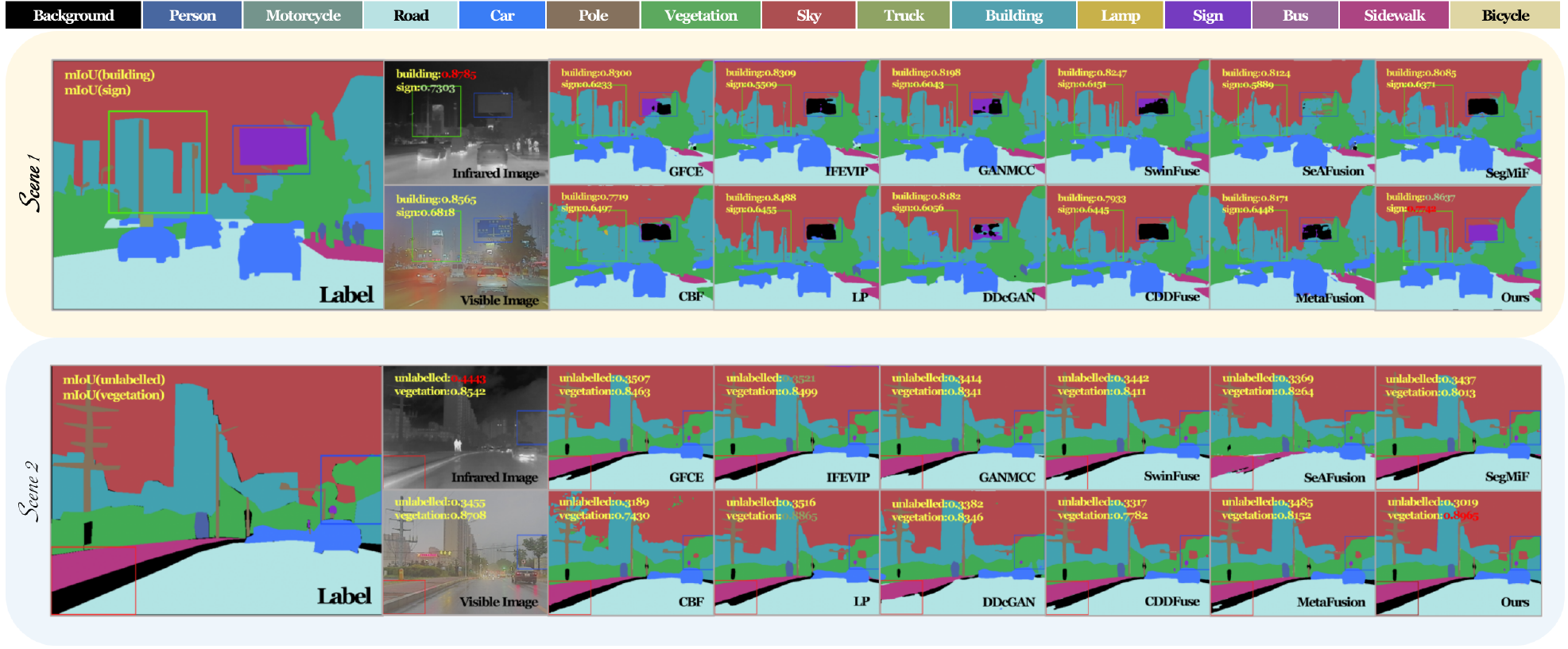}
\caption{Semantic segmentation results of source images, compared methods and HSFusion on FMB dataset.} \label{fig:fig6}
\end{figure}

\begin{table}
\centering
\caption{Semantic segmentation performance for comparison experiments on FMB dataset, where \textcolor{red}{red} values are optimal, \textcolor{green!70!black}{green} values are sub-optimal.}
\label{segmentation_result}
\resizebox{\textwidth}{!}{%
\begin{tabular}{@{}cccccccccccccccc@{}}
\toprule
Methods &  unlabelled & road & sidewalk & building & lamp & sign & vegetation & sky & person & car & truck & bus & motorcycle & pole & mIoU\\ \midrule
Visible & 0.3455 & 0.9020 & 0.7005 & 0.8565 & \textcolor{green!70!black}{0.5336} & 0.6818 & 0.8708 & 0.9379 & 0.6456 & 0.8355 & 0.6867 & 0.7715 & 0.5826 & 0.5416 & 0.7066 \\
Infrared & \textcolor{red}{0.4443} & 0.8985 & 0.6212 & \textcolor{red}{0.8785} & \textcolor{red}{0.6392} & \textcolor{green!70!black}{0.7303} & 0.8542 & \textcolor{green!70!black}{0.9412} & 0.7038 & \textcolor{green!70!black}{0.8497} & 0.4198 & 0.7513 & 0.4032 & 0.4977 & 0.6881 \\
CBF & 0.3189 & 0.9074 & 0.7312 & 0.7719 & 0.4808 & 0.6497 & 0.7430 & 0.7399 & 0.7172 & 0.8330 & 0.8295 & 0.8313 & 0.6276 & 0.5199 & 0.6930 \\
DDcGAN & 0.3382 & 0.8997 & 0.7102 & 0.8182 & 0.4451 & 0.6056 & 0.8346 & 0.8735 & 0.7092 & 0.8206 & 0.8101 & 0.8285 & 0.5859 & 0.5271 & 0.7005 \\
SeAFusion & 0.3369 & 0.9055 & 0.7237 & 0.8124 & 0.4481 & 0.5889 & 0.8264 & 0.8615 & 0.7132 & 0.8295 & 0.8208 & 0.8312 & 0.6187 & 0.5317 & 0.7035 \\
SwinFuse & 0.3442 & 0.9018 & 0.7151 & 0.8247 & 0.4571 & 0.6151 & 0.8411 & 0.8806 & 0.7130 & 0.8257 & 0.8171 & 0.8305 & 0.5997 & 0.5336 & 0.7071 \\
CDDFuse & 0.3317 & 0.9114 & 0.7473 & 0.7933 & 0.4792 & 0.6445 & 0.7782 & 0.7911 & 0.7272 & 0.8385 & 0.8328 & 0.8467 & 0.6479 & 0.5313 & 0.7072 \\
GANMCC & 0.3414 & 0.9064 & 0.7273 & 0.8198 & 0.4530 & 0.6043 & 0.8341 & 0.8712 & 0.7161 & 0.8327 & 0.8257 & 0.8351 & 0.6202 & 0.5375 & 0.7089 \\
IFEVIP &\textcolor{green!70!black}{0.3521} & 0.9043 & 0.7267 & 0.8309 & 0.4656 & 0.5509 & 0.8499 & 0.8876 & 0.7201 & 0.8302 & 0.8212 & 0.8401 & 0.6101 & 0.5375 & 0.7091 \\
GFCE & 0.3507 & 0.9042 & 0.7224 & 0.8300 & 0.4657 & 0.6233 & 0.8463 & 0.8856 & 0.7185 & 0.8301 & 0.8249 & 0.8383 & 0.6097 & 0.5381 & 0.7134 \\
LP & 0.3516 & 0.9040 & 0.7180 & 0.8488 & 0.4902 & 0.6455 & \textcolor{green!70!black}{0.8865} & 0.9398 & 0.7005 & 0.8361 & 0.7326 & 0.8473 & 0.5435 & \textcolor{green!70!black}{0.5518} & 0.7140 \\
SegMiF & 0.3437 & \textcolor{red}{0.9134} & \textcolor{red}{0.7558} & 0.8085 & 0.4882 & 0.6371 & 0.8013 & 0.8226 & \textcolor{red}{0.7355} & 0.8442 & 0.8417 & \textcolor{green!70!black}{0.8599} & \textcolor{red}{0.6637} & 0.5406 & 0.7183 \\
MetaFusion & 0.3485 & \textcolor{green!70!black}{0.9118} & \textcolor{green!70!black}{0.7528} & 0.8171 & 0.4890 & 0.6448 & 0.8152 & 0.8426 & \textcolor{green!70!black}{0.7318} & 0.8431 & \textcolor{green!70!black}{0.8451} & 0.8513 & \textcolor{green!70!black}{0.6624} & 0.5448 & \textcolor{green!70!black}{0.7215} \\
\textbf{HSFusion(Ours)} & 0.3019 & 0.9067 & 0.7512 & \textcolor{green!70!black}{0.8637} & 0.5043 & \textcolor{red}{0.7742} & \textcolor{red}{0.8965} & \textcolor{red}{0.9415} & 0.7235 & \textcolor{red}{0.8557} & \textcolor{red}{0.8678} & \textcolor{red}{0.8773} & 0.6599 & \textcolor{red}{0.5591} & \textcolor{red}{0.7488} \\
\bottomrule
\end{tabular}%
}
\end{table}
\subsection{Segmentation comparison and analysis}
\subsubsection{Qualitative comparison and analysis.}
To demonstrate the effectiveness of our method in high-level vision task, we compare it with eleven fusion methods in the semantic segmentitation task.  Fig.~\ref{fig:fig6}  shows the segmentation results of all the methods. The scene 1 of  Fig.~\ref{fig:fig6} is captured in nighttime,  which shows that our method can accurately segment the sign in the blue box, but other methods partially labeled it as unlabelled. Besides, for the building in the green box, our method can capture the structure of the building, thus achieving the highest segmentation $mIoU$ than other methods. In addition, the scene 2 of  Fig.~\ref{fig:fig6} is captured in daytime, and our results can still accurately segment the road and unlabelled labels in the red box. Meanwhile, our method achieves a fine distinction between vegetation and sky in the blue box. These advantages primarily stem from two factors. On the one hand, our fusion results preserve clearer structural information and texture details, thereby enhancing the segmentation performance. On the other hand, our semantic reinforce process successfully inject semantic information into the features that used in fusion process, which can enrich the final fusion results with meaningful semantic expression.
\subsubsection{Quantitative comparison and analysis.}
The results of quantitative comparison are illustrated in Table~\ref{segmentation_result}. The $IoU$ of each label indicates that our method has the best performance in most cases, such as the foreground( car, truck and bus) and background( building, sign, vegetation, sky and pole). Moreover, the $mIoU$ of our method is the highest among all methods, which demonstrates that compared with other methods, our fused results can significantly improve the performance in semantic segmentation tasks.
\section{Conclusion}
\label{sec:conclusion}
In this paper, to enhance the vision perception of the fused results and improve the performance in  high-level vision task. We propose a high-level vision task-driven infrared and visible image fusion network via semantic and geometric domain transformation, which contains two separate domain transformation branches to minimize the gap between semantic and geometric representation.  In fusion stage, we utilize the middle features of two separate CycleGANs to force the fused image capture more semantic information, which further improve the performance of our results in high level vision tasks. The comparative experiments demonstrate the superiority of our method both on visual appeal and high-level semantic segmentation tasks.

\end{document}